\documentclass[letterpaper]{article}
\usepackage{aaai}
\usepackage{helvet}
\usepackage{local}

\usepackage{url}

\usepackage[english, linesnumbered]{algorithm2e}
\SetKwBlock{Deb}{begin}{end}
\SetKwFor{ForEach}{for each}{do}{end}

\usepackage{graphicx}

\begin{document}

\title{
    Belief revision in the propositional closure of a qualitative algebra\thanks{
        This research was partially funded by the project Kolflow
        (\texttt{http://kolflow.univ-nantes.fr})
        of the French National Agency for Research (ANR), program ANR CONTINT.
    }
}
\author{%
  Valmi Dufour-Lussier$^{1, 2, 3}$ \and
  Alice Hermann$^{1, 2, 3}$ \and
  Florence Le~Ber$^{4}$ \and
  Jean Lieber$^{1, 2, 3}$
  \\
  \begin{tabular}{l}
  $^1$
  Universit\'e de Lorraine, LORIA, UMR 7503 --- 54506 Vand\oe{}uvre-l\`es-Nancy, France \\
  ~~~~\texttt{FirstName.LastName@loria.fr}
  \\
  $^2$
  CNRS --- 54506 Vand\oe{}uvre-l\`es-Nancy, France
  \\
  $^3$
    Inria --- 54602 Villers-l\`es-Nancy, France
  \\
  $^4$
  ICube -- Universit\'e de Strasbourg/ENGEES, CNRS --- 67412 Illkirch, France \\
  ~~~~\texttt{florence.leber@engees.unistra.fr}
  \end{tabular}}
\maketitle

\begin{abstract}
  Belief revision is an operation that aims at modifying old beliefs so that they
   become consistent with new ones.
  The issue of belief revision has been studied in various formalisms, in particular,
   in qualitative algebras (QAs) in which the result is a disjunction of belief bases
   that is not necessarily representable in a QA.
  This motivates the study of belief revision in formalisms extending QAs,
   namely, their propositional closures:
   in such a closure, the result of belief revision belongs
   to the formalism.
  Moreover, this makes it possible to define a contraction operator thanks to the Harper
   identity.
  Belief revision in the propositional closure of QAs is studied, an algorithm
   for a family of revision operators is designed, and
   an open-source implementation is made freely available on the web.

  \begin{description}
  \item[Keywords:]
        qualitative algebras,
        belief revision,
        belief contraction,
        propositional closure
  \end{description}
\end{abstract}

\section{Introduction}

Belief revision is an operation of belief change that consists
 in modifying minimally old beliefs so that they become consistent
 with new beliefs~\cite{agm-85}.
One way to study this issue following a knowledge representation angle
 is to consider a formalism and to study some belief revision operators
 defined on it:
 how they are defined and how they can be implemented.

In particular, it is rather simple to define a revision operator on a
 qualitative algebra (QA, such as the Allen algebra) by
 reusing the work of~\cite{condotta10} about the related issue of
 belief merging.
The result of such a belief revision is a set of belief bases to
 be interpreted disjunctively, and which is not necessarily representable
 as a single belief base:
 QAs are not closed under disjunction.

This gives a first motivation for the study of belief revision in the
 propositional closure of a QA:
 the revision operator in such a closure gives a result that is necessarily
 representable in the formalism.
Another motivation lies in the possibility of defining a contraction operator
 in this formalism, thanks to the Harper identity.

The first section of the paper contains some preliminaries about various
 notions used throughout the paper.
The next section briefly describes some properties of such a formalism.
Finally, an algorithm and an implementation of this algorithm for a revision
 operator in the propositional closure of a QA are
 presented with an example.

The research report~\cite{dufour2014_REVISOR_PCQA_long_version}
 is a long version of this paper including more detailed preliminaries,
 the proofs and some additional examples.

\section{Preliminaries}
\subsection{Qualitative algebras}

Qualitative algebras (QAs) are formalisms that are widely used
 for representation depending on time and/or on space~\cite{stock97}.
Formulas built upon QAs are closed under conjunction,
 though the symbol $\land$ is not systematically used.
Some of the usual notations and conventions of QAs
 are changed to better fit the scope of this paper.
In particular, the representation of knowledge by graphs
 (namely, qualitative constraint networks)
 is not well-suited here, because of the propositional closure
 introduced afterwards.

First, the Allen algebra is introduced: 
 it is one of the most famous QAs and it will be used in
 our examples throughout the paper.
Then, a general definition of QAs is given.

\subsubsection{The Allen algebra}
 is used for representing relations between time intervals~\cite{allen83cacm}.
A formula of the Allen Algebra can be seen as a conjunction of constraints, where
 a constraint is an expression of the form
 $\varx\relAQ\vary$ stating that the interval $\varx$ is related to
 the interval $\vary$ by the relation $\relAQ$.
$13$ base relations are introduced (cf. figure~\ref{fig:Allen});
 a relation $\relAQ$ is either one of these base relations
 or the union of base relations $\relAQNum1$, \ldots, $\relAQNum{m}$
 denoted by ${\relAQNum1}\ourel\ldots\ourel{\relAQNum{m}}$.

For example, if one wants to express that the maths course
 is immediately before the physics course
 which is before the English course
 (either with a time lapse, or immediately before it), one can write the
 formula:
 \begin{align*}
   \maths \mAQ \physique
   \quad\land\quad
   \physique \mathrel{{\bAQ}\ourel{\mAQ}} \anglais
 \end{align*}

$\logiqueAllen$ is the set of the formulas of the Allen algebra.

\begin{figure}
    \begin{center}
        {
  \def\intx{$\rule[1.5mm]{20mm}{1.5mm}$}
  \def\inty#1#2{{\textcolor{gray!50}{\hspace{#1mm}$\rule[0mm]{#2mm}{1.5mm}$}}}
  \def\ligne#1#2#3#4{\makebox[0mm][l]{\intx}\inty{#1}{#2} & $\nomrelAQ{#3}$ & \emph{#4}}
  \def\rien#1{#1} 
  %
    \begin{tabular}{l c l}
      \ligne{25}{15}{b}{is \rien{b}efore}
      \\[1.5mm]
      \ligne{20}{15}{m}{\rien{m}eets}
      \\[1.5mm]
      \ligne{15}{15}{o}{\rien{o}verlaps}
      \\[1.5mm]
      \ligne{0}{25}{s}{\rien{s}tarts}
      \\[1.5mm]
      \ligne{-5}{30}{d}{is \rien{d}uring}
      \\[1.5mm]
      \ligne{-5}{25}{f}{\rien{f}inishes}
      \\[1.5mm]
      \ligne{0}{20}{eq}{\rien{eq}uals}
    \end{tabular}
  %
}

    \end{center}
        $\biAQ$, $\miAQ$, $\oiAQ$, $\siAQ$, $\diAQ$ and $\fiAQ$
        represent respectively the inverse relations of
        the relations represented by $\bAQ$, $\mAQ$, $\oAQ$, $\sAQ$, $\dAQ$ and $\fAQ$.
  \caption{The base relations of $\logiqueAllen$.\label{fig:Allen}}
\end{figure}

\subsubsection{Qualitative algebras}
 in general are defined below, first by their syntax and then by their semantics.

\paragraph{Syntax.}
A finite set of symbols $\RelationsAQBases$ is given
 (with $|\RelationsAQBases|\geq2$).
A \emph{base relation} is an element of $\RelationsAQBases$.
A \emph{relation} is an expression of the
 form ${\relAQNum1}\ourel\ldots\ourel{\relAQNum{m}}$
 ($m\geq0$),
 such that a base relation occurs at most once in a relation
 and the order is irrelevant
 (e.g. ${\relAQNum1}\ourel{\relAQNum2}$ and ${\relAQNum2}\ourel{\relAQNum1}$
  are equivalent expressions).
The set of relations is denoted by $\RelationsAQ$.

A finite set of symbols $\Variables$, disjoint from $\RelationsAQBases$, is given.
A \emph{(qualitative) variable} is an element of $\Variables$.

A \emph{constraint} is an expression of the form $\varx\relAQ\vary$
 where $\varx, \vary\in\Variables$ and ${\relAQ}\in\RelationsAQ$.

A \emph{formula} $\varphi$ is a conjunction of $n$ constraints ($n\geq1$):
 $\varxNum1\relAQNum1\varyNum1\;\land\;\ldots\;\land\;\varxNum{n}\relAQNum{n}\varyNum{n}$.
A constraint of $\varphi$ is one of the constraints of this conjunction.
Let $\logiqueQA$ be the set of the formulas of the considered QA.
The atoms of $\logiqueQA$ are the constraints.

A formula $\varphi\in\logiqueQA$ is under normal form if
 for every $\varx, \vary\in\Variables$ with $\varx\neq\vary$,
 there is exactly one ${\relAQ}\in\RelationsAQ$ such that
 $\varx\relAQ\vary$ is a constraint of $\varphi$.
Then, this relation $\relAQ$ is denoted by $\relDe{\varphi}(\varx, \vary)$.

A \emph{scenario} $\scenario$ is a formula under normal form such that,
 for every variables $\varx$ and $\vary$, $\varx\neq\vary$,
 $\relDe{\scenario}(\varx, \vary)\in\RelationsAQBases$.
Therefore, there are $|\RelationsAQBases|^{|\Variables|\times(|\Variables|-1)}$
 scenarios.

\paragraph{Semantics.}
%
%
The semantics of a QA can be defined classically, thanks to a domain,
 a variable being mapped into a subset of this domain and a relation
 being mapped on a relation between such subsets.
For Allen algebra, the domain is the set $\Rationnels$ of rational numbers
 and, given an interpretation $\interpretation$,
 a variable $\varx$ is mapped to an interval $\interpretation(x)=[a, b]$
 of $\Rationnels$ ($a < b$).
The semantics of each of the basic relations is defined.
For example, $\interpretation$ satisfies $\varx_1\mAQ\varx_2$
 if $b_1=a_2$ where $\interpretation(\varx_i)=[a_i, b_i]$.
$\interpretation$ satisfies
 $\varx_1\mathrel{({\relAQNum1}\ourel\ldots\ourel{\relAQNum{m}})}\varx_2$
 if it satisfies one of the constraints $\varx_1\mathrel{\relAQNum{k}}\varx_2$
 for $k\in\{1, \ldots, m\}$.
A formula is consistent (or satisfiable) if there exists an interpretation
 satisfying each of its constraints.
Finally, for $\varphi_1, \varphi_2\in\logiqueQA$,
 $\varphi_1\models\varphi_2$ if,
 for every interpretation $\interpretation$ satisfying $\varphi_1$,
 it satisfies also $\varphi_2$.
The research report give more details on this first definition of the semantics.

The semantics can be characterized a posteriori thanks to
 consistent scenarios.

Let $\INTERPRETATIONS$ be the set of consistent scenarios on the
 variables of $\Variables$.
It can be 
 proven that
 $|\INTERPRETATIONS|\leq|\RelationsAQBases|^{|\Variables|\times(|\Variables|-1)/2}$:
 if $\varx\relAQ\vary$ is a constraint of a consistent scenario $\scenario$
 then $\vary\inverserel{\relAQ}\varx$ is also a constraint of $\scenario$.

Let $\Mod : \logique\rightarrow\EnsembleParties{\INTERPRETATIONS}$ be defined
 by
 \begin{equation*}
   \Mod(\varphi)=\{\scenario\in\INTERPRETATIONS~|~\scenario\models\varphi\}
 \end{equation*}
 for $\varphi\in\logique$, where $\models$ is the entailment relation
 defined below, thanks to the semantics based on a domain.

$\Omega$ and $\Mod$ make it possible to define a semantics on $\logique$
 which coincides with the semantics based on a domain
 (hence the same entailment relation $\models$):
 $\varphi_1\models\varphi_2$ iff $\Mod(\varphi_1)\subseteq\Mod(\varphi_2)$.
However, this second semantics is more practical to use for
 defining revision operators on QAs.

\subsection{Belief change}

\subsubsection{Belief revision}
 is an operation of belief change.
Intuitively, given the set of beliefs $\psi$ 
 an agent has about a static world, it consists in considering the
 change of their beliefs when faced with a new set of beliefs
 $\mu$, assuming that $\mu$ is considered to be unquestionable
 by the agent.
The resulting set of beliefs is noted $\psi\rev\mu$, and depends on the
 choice of a belief revision operator $\rev$.
In~\cite{agm-85}, the principle of minimal change has been stated
 and could be formulated as follows:
 $\psi$ is minimally changed into $\psi'$ such that the conjunction
 of $\psi'$ and $\mu$ is consistent,
 and the result of the revision is this conjunction.
Hence, there is more than one possible $\rev$ operator, since
 the definition of $\rev$ depends on how belief change is ``measured''.
More precisely, the minimal change principle has been formalized by
 a set of postulates, known as the AGM~postulates
 (after the names of the authors of~\cite{agm-85}).

In~\cite{katsuno91}, revision has been studied in the framework of
 propositional logic (with a finite set of variables).
The AGM~postulates are translated into this formalism
 and
 a family of revision operators is defined based on distance functions
 $\dist$ on $\INTERPRETATIONS$, where $\INTERPRETATIONS$ is the set of interpretations:
 the revision of $\psi$ by $\mu$ according to $\revDist$
 ($\psi\revDist\mu$) is such that
 \begin{align}
   \Mod(\psi\revDist\mu)
   &=
   \{\interpretationOmega\in\Mod(\mu)
    ~|~
    \dist(\Mod(\psi), \interpretationOmega) = \distPsiMu\}
    \notag
    \\
    \text{with }\distPsiMu &= \dist(\Mod(\psi), \Mod(\mu))
    \label{eq:revDist}
 \end{align}
Intuitively, $\distPsiMu$ measures, using $\dist$, the minimal modification
 of $\psi$ into $\psi'$ needed to make $\psi'\land\mu$ consistent.


This approach can be extended to other formalisms for which a model-theoretic
 semantics can be defined and such that a distance function can be specified on
 the set of interpretations $\INTERPRETATIONS$.
However, in some of these formalisms, a representability issue can be raised:
 it may occur that a subset $\SEI$ of $\INTERPRETATIONS$ is not representable,
 i.e. there is no formula $\varphi$ such that $\Mod(\varphi)=\SEI$.
This representability issue is addressed below, for the case of QAs.


\subsubsection{Belief contraction}
 is the operation of belief change that associates to a set of beliefs
 $\psi$ and a set of beliefs $\mu$, a set of beliefs $\psi\contraction\mu$
 such that $\psi\contraction\mu\not\models\mu$.
In propositionally closed formalisms, the Harper identity makes it possible to define
 a contraction operator $\contraction$ thanks to a revision operator $\rev$
 with
 \begin{equation}
   \psi\contraction\mu = \psi\lor(\psi\rev\lnot\mu)
   \label{eq:identite-Harper}
 \end{equation}

\subsubsection{Belief merging}
 is another operation of belief change.
Given some sets of beliefs $\psi_1$, \ldots, $\psi_n$,
 their merging is a set of beliefs
 $\Psi$ that contains ``as much as possible'' of the beliefs in
 the $\psi_i$'s.
Intuitively, $\Psi$ is the conjunction of $\psi_1'$, \ldots, $\psi_n'$
 such that each $\psi_i$ has been minimally modified into $\psi_i'$
 in order to make this conjunction consistent.
Some postulates of belief merging have been proposed and
 discussed~\cite{konieczny02},
 in a similar way as the AGM~postulates.


\subsection{Belief revision in qualitative algebras}


In~\cite{condotta10} a belief merging operator is defined
 that can be easily adapted for defining a revision operator.
It is based on a distance between scenarios.
Let $\distRel$ be a distance function on $\RelationsAQBases$.
Let $\scenario, \tau\in\INTERPRETATIONS$, be two scenarios based
 on the same set of variables $\Variables$.
Then, $\dist$ is defined by
 \begin{equation*}
   \dist(\scenario, \tau)
   =
   \sum_{\varx,\vary\in\Variables, \varx\neq\vary}
   \distRel(\relDe{\scenario}(\varx, \vary), \relDe{\tau}(\varx, \vary))
 \end{equation*}
One of the possibilities for $\distRel$ is the use of a neighborhood
 graph, i.e. a connected, undirected graph whose vertices are
 the base relations: $\distRel({\relAQr}, {\relAQs})$
 is the length of the shortest path between $\relAQr$ and $\relAQs$.
Then, the scenarios of the revision of $\psi$ by $\mu$ according to
 $\revDist$ are the scenarios of $\mu$ that are the closest ones to
 scenarios of $\psi$ according to $\dist$.
The set of optimal scenarios is not necessarily representable in
 $(\logiqueQA, \models)$.
One solution to address this issue 
 is to consider that the result of revision
 is a set of scenarios.


Algorithms for implementing this kind of belief revision in QAs are
 presented in~\cite{dufourlussier:hal-00735231} and~\cite{hue12revising}.

\section{Propositional closure of a qualitative algebra}

The propositional closure of a QA $(\logiqueQA, \models)$
 is a formalism $(\logiqueQACP, \models)$ defined as follows.
$\logiqueQACP$ is the smallest superset of $\logiqueQA$ that
 is closed for $\lnot$ and $\land$.
Then, $\logiqueQACP$ is closed for $\lor$:
 $\varphi_1\lor\varphi_2$ is an abbreviation for
 $\lnot(\lnot\varphi_1\land\lnot\varphi_2)$.
The entailment relation is based on the consistent scenarios:
 $\varphi_1\models\varphi_2$ if $\Mod(\varphi_1)\subseteq\Mod(\varphi_2)$,
 $\Mod$ being extended on $\logiqueQACP$ by
 $\Mod(\varphi_1\land\varphi_2)=\Mod(\varphi_1)\cap\Mod(\varphi_2)$
 and $\Mod(\lnot\varphi)=\INTERPRETATIONS\setminus\Mod(\varphi)$.

\begin{proposition}[representability]\label{prop:representabilite}
  Every set of scenarios $\SEI\subseteq\INTERPRETATIONS$
   is representable in $\logiqueQACP$.
  More precisely, with $\displaystyle\varphi=\bigvee_{\scenario\in\SEI}\scenario$,
   $\Mod(\varphi)=\SEI$.
\end{proposition}
%

Every formula of $\logiqueQACP$ can be written in DNF (disjunctive normal
 form, i.e. disjunction of conjunctions of constraints), since
 it is a propositionally closed formalism, but the following
 proposition goes beyond that.

\begin{proposition}[\DNFWoN form]\label{prop:formesNormales}
  Every $\varphi\in\logiqueQACP$ is equivalent to a formula in
   DNF using no negation symbol.
\end{proposition}
\section{Belief revision in $(\logiqueQACP, \models)$}

Given a distance function $\dist$ on $\INTERPRETATIONS$,
 a revision operator on $(\logiqueQACP, \models)$ can be
 defined according to equation~(\ref{eq:revDist}).
Indeed, proposition~\ref{prop:representabilite} implies
 that
 $\{\interpretationOmega\in\Mod(\mu)~|~\dist(\Mod(\psi), \interpretationOmega) = \distPsiMu\}$
 is representable.

\subsection{An algorithm for computing $\revDist$ in $\logiqueQACP$}

The principle of the algorithm is based on the following proposition.

\begin{proposition}[revision of disjunctions]\label{prop:revision-disjonctions}
  Let $\psi$ and $\mu$ be two formulas of $\logiqueQACP$ and
   $\{\psi_i\}_i$ and $\{\mu_j\}_j$ be two finite
   families of $\logiqueQACP$
   such that
   $\displaystyle\psi=\bigvee_i\psi_i$ and
   $\displaystyle\mu=\bigvee_j\mu_j$.

  Let 
   $\distPsiMu_{ij}=\dist(\Mod(\psi_i), \Mod(\mu_j))$
   for any $i$ and $j$.
  Then:
  \begin{align}
    \psi\revDist\mu
    &\equiv
    \bigvee_{i, j, \distPsiMu_{ij}=\distPsiMu}
    \psi_i\revDist\mu_j
    \notag
    \\
    \text{with }
    \distPsiMu
    &=
    \dist(\Mod(\psi), \Mod(\mu))
    \notag
    \\
    \text{Moreover,}\quad
    \distPsiMu
    &=
    \min_{ij}\distPsiMu_{ij}
    \label{eq:Delta=minDeltaij}
  \end{align}
\end{proposition}

The algorithm for $\revDist$ in $\logiqueQACP$ consists
 roughly in putting $\psi$ and $\mu$ in $\DNFWoN$ form
 then applying proposition~\ref{prop:revision-disjonctions}
 on them, using the $\revDist$ algorithm on $\logiqueQA$
 for computing the $\psi_i\revDist\mu_j$'s.
More details are given in the research report.


\subsection{Implementation: the \revisorPCQA engine}

\revisor
 is a collection of several revision engines that are open-source and
 freely available.\footnote{\url{http://revisor.loria.fr}}

In particular, \revisorQA implements $\revDist$ in three QAs:
 the Allen algebra,
 INDU---an extension of the Allen algebra taking into account relations between intervals
 according to their lengths~\cite{pujari99indu}---and
 RCC8---a QA for representing topological relations between regions of
 space~\cite{randell92spatial}.
Moreover, it is easy to use a different qualitative algebra,
 by giving in the code some tables
 (composition table, inverse relation table,
  and table for the values $\distRel(\relAQr, \relAQs)$ for $\relAQr,
  \relAQs\in\RelationsAQBases$).
The engine is written in Perl,
 but can be used through a Java library.
The worst-case complexity of this implementation is of the order of
 $O\left(|\RelationsAQBases|^{\frac{|\Variables|\cdot(|\Variables|-1)}{2}}\right)$.

\revisorPCQA
 implements $\revDist$ on the propositional closures of the QAs $\logiqueAllen$,
 INDU and RCC8:
 it actually uses \revisorQA
 and is one of the engines of \revisor.
The worst-case complexity of this implementation is of the order of
 $O\left(|\Variables|^4|\RelationsAQBases|^{\frac{|\Variables|\cdot(|\Variables|-1)}{2}}\right)$,
 according to a coarse analysis.

%
The following example has been executed using \revisorPCQA,
 and is included with the source code.
The \verb=README= file associated with \revisorQA on the \revisor
 website explains how it can be executed.



\def\Boole{\fm{Boole}}
\def\DeMorgan{\fm{De~Morgan}}
\def\Weierstrass{\fm{Weierstra\ss}}
%
%
This example
 uses a belief contraction operator.
According to~(\ref{eq:identite-Harper}),
 a contraction operator $\contractionDist$ can be defined that is based on $\revDist$.
Now let us consider the set of beliefs $\psi$ of an agent called Maurice
 about the dates of birth and death of famous mathematicians.
Maurice thought that Boole was born after de~Morgan and died before
 him and that de~Morgan and Weierstra{\ss} were born the same year
 (say, at the same time) but the former died before the latter:
 \begin{align*}
   \psi
   =
   \Boole\dAQ\DeMorgan
   \;\land\;
   \DeMorgan\sAQ\Weierstrass
 \end{align*}
 where, $\Boole$ is the interval of time between the birth and
 the death of Boole, and so on.
Now, Germaine, a friend of Maurice, tells him that she is not sure whether
 Boole was born strictly after Weierstra{\ss}.
Since Maurice trusts Germaine (and her doubts), he wants to make the contraction of
 its original beliefs $\psi$ by $\mu$ with
 \begin{align*}
   \mu
   =
   \Boole
   \mathrel{{\biAQ} \ourel {\miAQ} \ourel {\oiAQ} \ourel {\fAQ} \ourel {\dAQ}}
   \Weierstrass
 \end{align*}
The result, computed by \revisorPCQA in less than one second, is
 $\psi\contractionDist\mu$, equivalent to the following formula:
 \begin{align*}
   &\strut\mathrel{\text{\phantom{$\lor$}}}
   (\Boole\dAQ\DeMorgan
    \;\land\;
    \DeMorgan\sAQ\Weierstrass)
    \\
   &\strut\lor
    (\Boole\sAQ\Weierstrass
     \;\land\;
     \DeMorgan\diAQ\Weierstrass)
     \\
   &\strut\lor
    (\Boole\sAQ\DeMorgan
     \;\land\;
     \DeMorgan\sAQ\Weierstrass
     )
     \\
   &\strut\lor
    \left(\!\!\!\!\mlc{$\Boole\dAQ\DeMorgan
                        \;\land\;
                        \Boole\sAQ\Weierstrass$ \\
                       $\;\land\;
                        \DeMorgan\oAQ\Weierstrass$}\!\!\!\!\right)
 \end{align*}
Actually, the last term of this disjunction corresponds to the
 reality, provided that the intervals of time correspond to a
 year granularity:
 George Boole (1815-1864),
 Augustus De~Morgan (1806-1871),
 Karl Weierstra{\ss} (1815-1897).

In~\cite{dufour2014_REVISOR_PCQA_long_version}, other examples
 of use of \revisorPCQA, including an analysis of the computing time,
 are presented.
In particular, it is shown that it may be the case that a family
 of revision problems, albeit formalizable in both $\logiqueAllen$
 and in $\logiqueAllenCP$, are solved in much less time in the more
 expressive formalism $\logiqueAllenCP$.

\section{Conclusion}

This paper has presented an algorithm for distance-based belief revision in the
 propositional closure $\logiqueQACP$ of a qualitative algebra $\logiqueQA$,
 using the revision operation on $\logiqueQA$. 
This work is motivated by the fact that it gives a revision operation whose
 result is representable in the formalism,
 by the fact that some practical examples are easily represented in $\logiqueQACP$
 whereas they are quite difficult to represent in $\logiqueQA$,
 and by the fact that it makes it possible to define a contraction operator thanks to the
 Harper identity (which requires disjunction and negation).
The preprocessing of the algorithm consists in putting the formulas into
 a disjunctive normal form without negation.
Then, proposition~\ref{prop:revision-disjonctions},
 which reduces a revision of disjunctions to a disjunction
 of the least costly revisions, is applied.
\revisorPCQA is an implementation of this revision operator for
 the Allen algebra, INDU and RCC8.

A first direction of research following this work is the improvement of the
 computation time of the \revisorPCQA system.
One way to do it is to parallelize it.
A sequential optimization would consist in finding a heuristic for
 ranking the pairs $(i, j)$, with the aim of starting from the best candidates,
 in order to obtain a low upper bound of $\distPsiMu$ sooner
 (hence a pruning of a part of the search trees developed for the
  computation of $\revDist$ in $(\logiqueQA, \models)$).


Another direction of research is to study how other belief change
 operations can be implemented in this formalism,
 in particular belief merging~\cite{konieczny02}
 and knowledge update~\cite{katsuno-mendelzon:1991a}.

\bibliography{biblio}

\end{document}